\documentclass{article}

\usepackage[T1]{fontenc}
\usepackage[utf8]{inputenc}
\usepackage{microtype}
\usepackage{graphicx}
\usepackage{subcaption}
\usepackage{booktabs}
\usepackage{hyperref}
\usepackage{algorithm}
\usepackage{algorithmic}

\usepackage[accepted]{icml2026}
\makeatletter
\renewcommand{\ICML@appearing}{\textit{Accepted at the 2026 Workshop on Generative and Agentic AI for Biology (ICML 2026). Published on 29 May 2026.}}
\makeatother

\usepackage{amsmath}
\usepackage{amssymb}
\usepackage{mathtools}
\usepackage{amsthm}
\usepackage{multirow}
\usepackage{pifont}
\usepackage{colortbl}
\usepackage[capitalize,noabbrev]{cleveref}

\icmltitlerunning{De Novo Generation of Odorant Molecules with Targeted OR Activation Patterns}

\begin{document}

\twocolumn[
  \icmltitle{De Novo Generation of Odorant Molecules \\
    with Targeted Olfactory Receptor Activation Patterns}

  \icmlsetsymbol{equal}{*}

  \begin{icmlauthorlist}
    \icmlauthor{Kexin Zhang}{cs}
    \icmlauthor{Manhao Guan}{cs,ai}
  \end{icmlauthorlist}

  \icmlaffiliation{cs}{College of Computer Science and Technology, Zhejiang University, Hangzhou, China}
  \icmlaffiliation{ai}{College of Artificial Intelligence, Zhejiang University, Hangzhou, China}
  \icmlcorrespondingauthor{Manhao Guan}{guanmh@zju.edu.cn}

  \icmlkeywords{molecular generation, reinforcement learning, olfactory receptors, combinatorial code}

  \vskip 0.3in
]

\printAffiliationsAndNotice{}

\begin{abstract}
Odor perception arises from a combinatorial code in which each molecule activates a specific subset of olfactory receptors (ORs), and the resulting activation pattern determines the perceived smell. We present a framework for de novo generation of molecules predicted to match target multi-receptor activation profiles. We first construct a panoramic binding matrix over 20 ORs, then cluster it to discover representative activation patterns, and finally formulate a structured selectivity loss, combining pairwise ranking, distribution matching, and complexity constraints, to steer a molecular generator via reinforcement learning. Across nine target profiles of varying complexity, our method achieves high predicted selectivity within 18{,}000 oracle calls and provides the largest gains on profiles that are rare under the prior, especially broad-spectrum and sparse-activator targets.
\end{abstract}

\section{Introduction}
\label{sec:intro}

The mammalian olfactory system employs a combinatorial coding strategy to discriminate among a vast chemical space of volatile molecules \cite{malnic1999combinatorial,kurian2021odor_coding}. Each odorant activates a characteristic subset of the approximately 400 human olfactory receptors (ORs), and the identity of a perceived smell is encoded not by any single receptor but by the ensemble activation pattern across the receptor repertoire \cite{buck2004olfactory,kurian2021odor_coding}. This combinatorial logic enables a limited receptor repertoire to encode an enormous diversity of odor percepts.

Despite growing interest in computational approaches to molecular design \cite{du2024mlgendesign}, most existing work on odorant generation has focused on single-target objectives: optimizing a molecule's predicted binding affinity to one receptor or its similarity to a known odorant \cite{sanchez2019machine_scent,lee2023principal}. Such formulations neglect the multi-receptor nature of odor coding. A molecule that strongly activates a single target receptor may simultaneously activate unintended receptors, producing an entirely different odor percept than intended.

We argue that rational odorant design should instead target the \emph{activation profile}, the full pattern of which receptors a molecule activates and which it does not. This reframes the problem from single-objective optimization to multi-receptor selectivity control, directly reflecting the biological mechanism of odor encoding.

To this end, we propose a three-stage framework (\cref{fig:pipeline}):

\begin{figure*}[htbp]
  \centering
  \includegraphics[width=\textwidth]{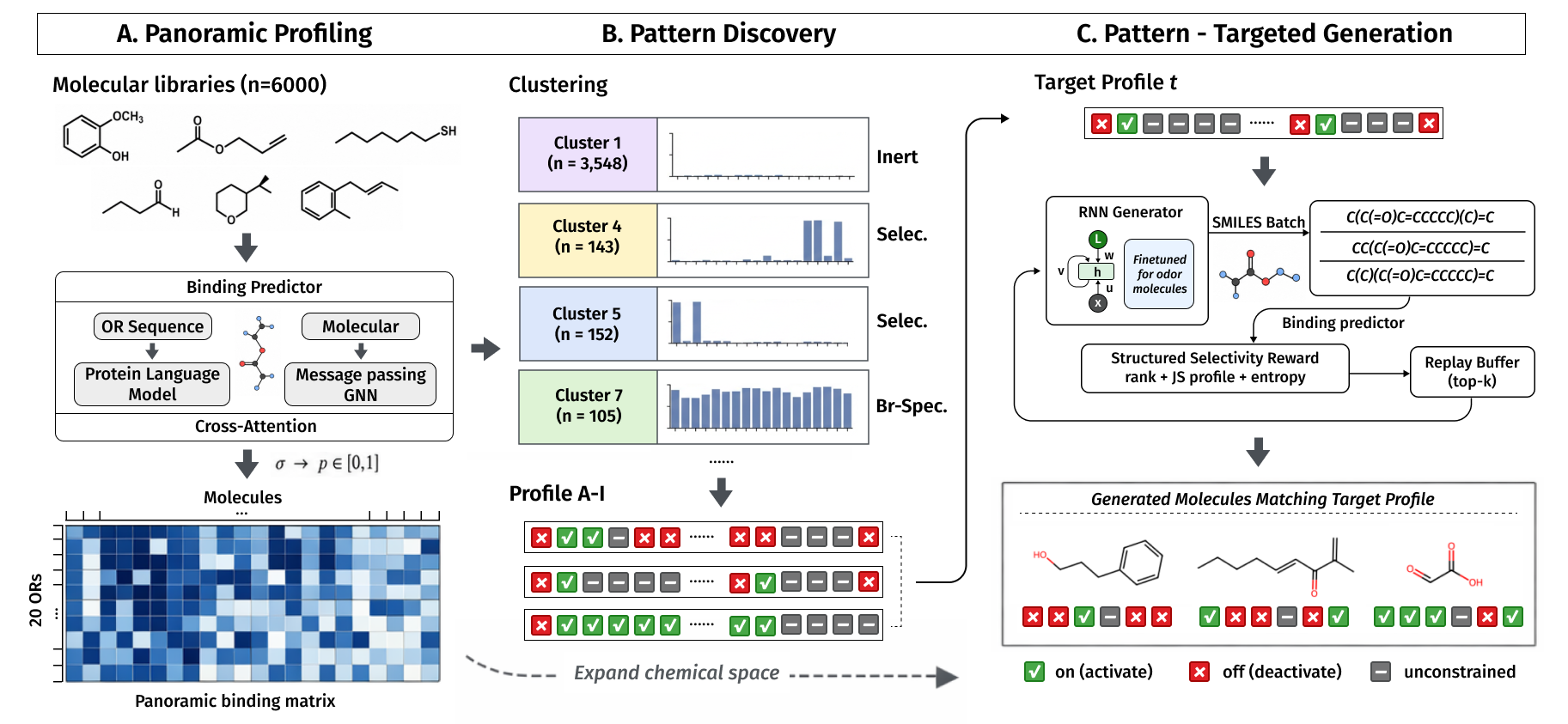}
  \caption{Overview of our framework. \textbf{Stage 1:} Panoramic profiling predicts binding probabilities of a molecular library against 20 ORs, producing a binding matrix. \textbf{Stage 2:} Clustering discovers representative activation patterns, interpreted as combinatorial receptor codes. \textbf{Stage 3:} A target pattern is selected and used to define a structured selectivity reward combining ON/OFF ranking, response-distribution matching, and activation-complexity matching. RL with augmented memory steers the generator toward molecules predicted to match the target profile.}
  \label{fig:pipeline}
\end{figure*}

The framework proceeds in three stages. We first perform \textbf{panoramic profiling} by predicting the binding probability of a diverse molecular library against a panel of 20 ORs, constructing a molecules $\times$ receptors binding matrix. We then apply \textbf{pattern discovery}, clustering this matrix to identify naturally occurring activation patterns shared by groups of molecules with similar multi-receptor profiles. Finally, we perform \textbf{pattern-targeted generation}, defining a selectivity reward that measures how well a generated molecule's predicted profile matches a chosen target pattern and using reinforcement learning (RL) to steer a molecular generator accordingly.

We evaluate our approach on nine target profiles of increasing complexity, from narrow selectivity over two receptors to broad-spectrum activation of fourteen. Our contributions are: (1) we formulate odorant generation as \emph{profile-targeted optimization} over a multi-receptor olfactory code rather than single-receptor binding; (2) we introduce a \emph{structured selectivity objective} combining ON/OFF ranking, response-distribution matching, and activation-complexity matching, providing a principled alternative to heuristic reward engineering; and (3) we evaluate RL-based molecular generation across nine diverse profiles and show that directed generation provides the largest gains for broad-spectrum and sparse-activator profiles, where screening from the prior distribution is insufficient, revealing a ``designability gradient'' relating profile complexity to optimization difficulty. We additionally explore a Lagrangian constrained variant that can accelerate simple selective profiles but becomes unstable for complex multi-constraint targets.


\section{Related Work}
\label{sec:related}

\paragraph{Olfactory receptor modeling.}
Predicting odorant--receptor interactions computationally has been approached through molecular docking \cite{charlier2012molecular}, learned perceptual representations \cite{sanchez2019machine_scent,shuvaev2024deepnose}, and graph neural networks operating on molecular structure \cite{hladis2023rec2odorant}. Recent work on odorant featurization has explored fine-tuning chemical foundation models to produce olfaction-specific representations \cite{mcconachie2025lorax}. The principal odor map \cite{lee2023principal} demonstrated that learned molecular embeddings can unify diverse olfactory tasks. The M2OR database \cite{lalis2024m2or} provides a large curated collection of OR--molecule interaction data, enabling supervised training of binding predictors across hundreds of receptors. Our work builds on these advances by using a trained binding predictor as an oracle within a generative loop, rather than as a standalone prediction tool.

\paragraph{De novo molecular generation.}
Deep generative models for molecules span variational autoencoders \cite{gomez2018automatic}, graph-based policy networks \cite{you2018gcpn}, and autoregressive models over SMILES strings \cite{segler2018generating}. Reinforcement learning approaches, particularly REINVENT \cite{olivecrona2017molecular,blaschke2020reinvent} and its Transformer-based extensions \cite{xu2024reinvent_transformer}, have shown strong performance in goal-directed generation by treating a pretrained language model as a policy and optimizing it against a scoring function. Recent advances in sample efficiency include augmented memory \cite{guo2024augmented}, which replays high-scoring molecules with diverse SMILES representations, and the Mamba-based Saturn architecture \cite{guo2026saturn}, which further improves oracle efficiency. Latent-space RL methods \cite{haddad2025molrl} offer an alternative by optimizing in continuous space. \citet{du2024mlgendesign} provide a comprehensive survey of machine learning-aided generative molecular design. We adopt the SMILES-based RL paradigm but extend the reward formulation from single-property optimization to multi-receptor profile matching.

\paragraph{Multi-objective molecular optimization.}
Prior work on multi-objective generation has addressed combinations of drug-likeness, synthetic accessibility, and target affinity \cite{jin2020multi,thomas2022augmented,guo2026tango}. However, optimizing a molecule's binding profile across a \emph{panel} of related biological targets, where some targets should be activated and others suppressed, has received limited attention. Known failure modes in molecular optimization \cite{renz2020failure} further motivate careful reward design. Our structured selectivity objective addresses this gap by combining profile-level matching with pairwise ranking between ON and OFF receptors, drawing on ideas from learning-to-rank \cite{liu2009learning} to enforce relative selectivity rather than relying solely on absolute binding predictions.

\section{Methods}
\label{sec:methods}

\subsection{Receptor--Odorant Binding Predictor}
\label{sec:binding_predictor}

We employ a dual-stream architecture to predict the binding probability between an odorant molecule and an OR sequence. The receptor stream encodes the OR amino acid sequence using a pretrained protein language model (ProtBERT; \citealp{elnaggar2021prottrans}), producing a fixed-dimensional receptor embedding. The molecular stream represents the odorant as a graph where atoms are nodes and bonds are edges, processed by a message-passing neural network (MPNN) with multi-head attention layers \cite{gilmer2017neural}. At each message-passing step, the receptor embedding is injected into every atom node via a cross-attention mechanism, enabling the network to learn receptor-specific molecular representations. After five rounds of message passing, a graph-level readout followed by a sigmoid output layer produces a binding probability $p \in [0, 1]$. The model is trained on the M2OR dataset \cite{lalis2024m2or}, which contains experimentally validated OR--molecule interactions across 566 receptors. We select 20 receptors with sufficient training data ($\geq 30$ positive interactions) for our panel. We denote the trained predictor as $f_r(m)$: the predicted binding probability of molecule $m$ to receptor $r$.

\paragraph{Predictor validation.} We evaluate the binding predictor on held-out annotated receptor--molecule pairs using a pair-level train/validation/test split. For each positive interaction in the held-out set, we sample three unannotated receptor--molecule pairs as negatives. Because OR interaction datasets are sparsely labeled, these random negatives may include unknown binders; thus the reported metrics should be interpreted as approximate surrogate-validation metrics rather than definitive binding performance. Across our 20-receptor panel, the predictor achieves a mean AUROC of $0.814 \pm 0.124$ and mean AUPRC of $0.654 \pm 0.137$. Performance varies by receptor: high-promiscuity receptors with many annotated interactions (e.g., s\_449, s\_51) show lower AUROC ($0.58$--$0.66$), likely due to the difficulty of distinguishing true binders from weakly interacting or unlabeled molecules, while specialist receptors (e.g., s\_1107, s\_177) achieve AUROC $> 0.91$. All selectivity scores reported in this work are \emph{predicted} values and require experimental validation.

\subsection{Molecular Generator}
\label{sec:generator}

Our molecular generator is a recurrent neural network (RNN) that autoregressively produces SMILES strings token by token \cite{segler2018generating}. The architecture consists of a multi-layer LSTM with a conditioning mechanism: at each decoding step, a receptor pocket embedding $\mathbf{z} \in \mathbb{R}^{d}$ is concatenated to the input, biasing generation toward molecules compatible with the target receptor context. The model is first pretrained on a large corpus of known odorant SMILES to learn the distribution of valid, chemically plausible odorant molecules, then fine-tuned with RL.


\subsection{Panoramic Profiling and Pattern Discovery}
\label{sec:panoramic}

To map the landscape of OR activation patterns, we sample 6{,}000 molecules from the pretrained generator, retain 5{,}816 valid molecules after RDKit parsing, and predict each valid molecule's binding probability against all 20 receptors, yielding a binding matrix $\mathbf{B} \in \mathbb{R}^{5816 \times 20}$ where $B_{ij} = f_{r_j}(m_i)$.

We cluster this matrix using $k$-means to identify $K$ representative activation patterns (receptor response prototypes) $\{\mathbf{c}_1, \ldots, \mathbf{c}_K\} \subset \mathbb{R}^{20}$. Each molecule's activation profile can be soft-assigned to prototypes via temperature-scaled cosine similarity:
\begin{equation}
    a_k(m) = \frac{\exp\!\left(\text{sim}(\mathbf{f}(m), \mathbf{c}_k) / \tau\right)}{\sum_{j=1}^{K} \exp\!\left(\text{sim}(\mathbf{f}(m), \mathbf{c}_j) / \tau\right)}
    \label{eq:soft_assign}
\end{equation}
where $\mathbf{f}(m) = [f_{r_1}(m), \ldots, f_{r_{20}}(m)]$ is the full binding profile and $\tau$ is a temperature parameter. These prototypes serve as a codebook of canonical activation patterns, enabling target specification via prototype selection or interpolation. In our experiments, we use the cluster centroids directly to define target profiles.

\subsection{Structured Combinatorial Code Loss}
\label{sec:reward}

We design a structured selectivity objective that captures three aspects of combinatorial receptor coding: relative selectivity, response geometry, and activation complexity.

\paragraph{Pairwise ranking selectivity.} The core selectivity requirement is that ON receptors should be activated \emph{more than} OFF receptors. We encode this via a pairwise margin-based ranking loss. For each (ON, OFF) receptor pair $(r^+, r^-)$:
\begin{equation}
    \mathcal{L}_{\text{rank}}(m) = \sum_{(r^+, r^-) \in \mathcal{P}} \log\!\left(1 + e^{-(f_{r^+}(m) - f_{r^-}(m) - \delta)}\right)
    \label{eq:ranking_loss}
\end{equation}
where $\mathcal{P}$ is the set of all ON--OFF pairs and $\delta = 0.3$ is a margin. This is converted to a reward $S_{\text{rank}} \in [0, 1]$ by normalizing against the maximum possible loss. Unlike absolute regression, this formulation depends on relative ordering and is robust to calibration error in the binding predictor.

\paragraph{Profile distribution matching.} We treat the receptor response as a distribution over the receptor panel and measure how well the generated molecule's response matches the target profile geometry. Using Jensen-Shannon divergence:
\begin{equation}
    S_{\text{dist}}(m) = 1 - \text{JSD}\!\left(\hat{\mathbf{p}}(m) \,\|\, \mathbf{p}_{\text{target}}\right)
    \label{eq:js}
\end{equation}
where $\text{JSD}$ denotes the Jensen-Shannon divergence computed with base-2 logarithms (so that $\text{JSD} \in [0,1]$), and $\hat{\mathbf{p}}(m)$, $\mathbf{p}_{\text{target}}$ are the $\ell_1$-normalized binding profiles. This captures the full response geometry rather than treating receptors as independent binary targets.

\paragraph{Activation complexity matching.} Different target profiles have different activation complexity: narrow profiles activate few receptors (low entropy) while broad profiles activate many (high entropy). We penalize deviation from the target's activation entropy:
\begin{equation}
    S_{\text{complex}}(m) = 1 - \left| H(\hat{\mathbf{p}}(m)) - H(\mathbf{p}_{\text{target}}) \right|
    \label{eq:complexity}
\end{equation}
where $H(\cdot)$ is the normalized Shannon entropy. This encourages the generator to match not just \emph{which} receptors are active, but the \emph{shape} of the combinatorial code.

\paragraph{Combined structured loss.} The final selectivity score combines all three components:
\begin{equation}
    S(m) = \alpha_{\text{rank}} \cdot S_{\text{rank}} + \alpha_{\text{dist}} \cdot S_{\text{dist}} + \alpha_{\text{complex}} \cdot S_{\text{complex}}
    \label{eq:combined}
\end{equation}
with $\alpha_{\text{rank}} = 0.4$, $\alpha_{\text{dist}} = 0.4$, $\alpha_{\text{complex}} = 0.2$. The reward incorporates quality gates:
\begin{equation}
    R(m) = S(m) \cdot \mathbb{1}[\text{OdorantScore}(m) \geq \tau_o] \cdot \mathbb{1}[\text{QED}(m) \geq \tau_q]
    \label{eq:reward}
\end{equation}
where $\tau_o = 0.5$ and $\tau_q = 0.3$.

\subsection{Reinforcement Learning with Augmented Memory}
\label{sec:rl}

We optimize the generator using the augmented memory framework \cite{guo2024augmented}, which extends the REINVENT algorithm \cite{olivecrona2017molecular}. The generator is duplicated into a frozen \emph{prior} $\pi_{\text{prior}}$ and a trainable \emph{agent} $\pi_\theta$. At each step, the agent samples a batch of SMILES strings conditioned on a pocket embedding $\mathbf{z}$, computed as the weighted mean of the pocket embeddings of the ON receptors in the target profile.

The loss for each generated molecule $m$ with SMILES representation $x$ is:
\begin{equation}
    \mathcal{L}(x) = \left( \log \pi_{\text{prior}}(x | \mathbf{z}) + \sigma \cdot R(m) - \log \pi_\theta(x | \mathbf{z}) \right)^2
    \label{eq:loss}
\end{equation}
where $\sigma$ controls the reward scaling. This loss drives the agent toward high-reward regions while maintaining proximity to the prior, preventing mode collapse.

The full training procedure is summarized in \cref{alg:selectivity_rl}.

\begin{algorithm}[htbp]
  \caption{Profile-Targeted Molecular Generation}
  \label{alg:selectivity_rl}
  \begin{algorithmic}
    \STATE {\bfseries Input:} Target profile $\mathbf{t}$, prototypes $\{\mathbf{c}_k\}$, predictor $f$, budget $N$
    \STATE Initialize prior $\pi_{\text{prior}}$, agent $\pi_\theta \leftarrow \pi_{\text{prior}}$
    \STATE Compute pocket embedding $\mathbf{z} \leftarrow \text{WeightedMean}(\{f.\text{embed}(r_i)\}_{t_i=1})$
    \STATE Initialize replay buffer $\mathcal{D} \leftarrow \emptyset$, calls $\leftarrow 0$
    \STATE Optionally initialize multipliers $\boldsymbol{\lambda}$ for constrained variant
    \WHILE{calls $< N$}
      \STATE Sample batch $\{x_1, \ldots, x_B\} \sim \pi_\theta(\cdot | \mathbf{z})$
      \STATE Compute $S_{\text{rank}}$, $S_{\text{dist}}$, $S_{\text{complex}}$ (\cref{eq:ranking_loss,eq:js,eq:complexity})
      \STATE $R(m_i) \leftarrow$ combined reward (\cref{eq:combined,eq:reward}); calls $\leftarrow$ calls $+ B$
      \IF{using constrained variant}
        \STATE Update $\boldsymbol{\lambda}$ via dual ascent (\cref{eq:lagrangian})
      \ENDIF
      \STATE Update $\mathcal{D}$ with top-$k$ molecules
      \STATE Compute $\mathcal{L}$ over batch $\cup$ replay samples (\cref{eq:loss})
      \STATE Update $\theta$ via gradient descent on $\mathcal{L}$
      \FOR{$j = 1$ {\bfseries to} $A$ (augmentation rounds)}
        \STATE Randomize SMILES of batch and replay samples
        \STATE Compute and backpropagate augmented loss
      \ENDFOR
    \ENDWHILE
    \STATE {\bfseries Return:} Trained agent $\pi_\theta$, top molecules from $\mathcal{D}$
  \end{algorithmic}
\end{algorithm}

\paragraph{Constrained optimization view.} The selectivity task can also be viewed as a constrained optimization problem: maximize ON receptor activation subject to OFF receptor suppression below a threshold $\epsilon$. We implement this via Lagrangian relaxation, maintaining dual variables $\lambda_j$ for each OFF receptor constraint:
\begin{equation}
    R_{\text{Lag}}(m) = \sum_{i \in \text{ON}} f_{r_i}(m) - \sum_{j \in \text{OFF}} \lambda_j \cdot \max(0, f_{r_j}(m) - \epsilon)
    \label{eq:lagrangian}
\end{equation}
The multipliers are updated via dual ascent after each batch: $\lambda_j \leftarrow \text{clip}(\lambda_j + \eta \cdot (\bar{f}_{r_j} - \epsilon), 0, \lambda_{\max})$, where $\bar{f}_{r_j}$ is the batch-mean OFF score. This automatically increases the penalty on violated constraints during training, providing an adaptive alternative to fixed reward weighting.

\section{Experimental Setup}
\label{sec:setup}

\paragraph{Receptor panel.} From the M2OR dataset \cite{lalis2024m2or}, we select 20 ORs with at least 30 positive interaction records each. Per-receptor binding rates in the panoramic matrix range from 0.9\% to 32\%, reflecting the natural heterogeneity of receptor promiscuity.

\paragraph{Target profiles.} We define nine target profiles spanning a range of complexity, derived from the cluster analysis of the panoramic binding matrix (\cref{tab:profiles}). \textbf{Profile A} (narrow selective) uses 2 ON / 3 OFF receptors, targeting the activation pattern of Cluster~5 while suppressing Cluster~4 receptors. \textbf{Profile B} (medium selective) uses 3 ON / 2 OFF receptors, targeting Cluster~6 against Cluster~4. \textbf{Profile C} (cross-cluster selective) uses 2 ON / 2 OFF receptors, targeting Cluster~0 while suppressing receptors dominant in Cluster~6. \textbf{Profile D} (broad-spectrum) uses 14 ON / 2 OFF receptors, targeting the broad activation pattern of Cluster~7. \textbf{Profile E} (anti-correlated) uses 2 ON / 2 OFF receptors (s\_449, s\_51 vs.\ s\_29, s\_443), two pairs of highly correlated receptors that are uncorrelated across pairs. \textbf{Profile F} (group vs.\ group) uses 3 ON / 3 OFF receptors, testing group-level selectivity between two receptor clusters. \textbf{Profile G} (mixed medium) uses 3 ON / 2 OFF receptors with moderate receptor promiscuity. \textbf{Profile H} (sparse activators) uses 3 ON / 2 OFF receptors targeting low-promiscuity receptors (s\_1116, s\_272, s\_464) with low baseline binding rates. \textbf{Profile I} (very broad) uses 9 ON / 3 OFF receptors, a high-dimensional activation target second only to Profile~D in complexity.

\paragraph{Training details.} For each profile, we run RL with a budget of 18{,}000 oracle calls (binding predictions), batch size 64, learning rate $10^{-4}$, reward scaling $\sigma = 60$, replay buffer size 100, and $A=2$ rounds of SMILES augmentation per step. Each run is repeated with 3 random seeds. All experiments use a single GPU and complete in approximately 15 minutes per run. For each seed, we report the peak batch-mean selectivity score achieved within the budget; \cref{tab:results} reports mean $\pm$ standard deviation over three seeds. This metric reflects the best average quality of a generated batch rather than a single best molecule.

\paragraph{Evaluation metrics.} Beyond selectivity score, we evaluate generated molecules on: validity (fraction parseable by RDKit), uniqueness (fraction of distinct canonical SMILES), QED \cite{bickerton2012quantifying}, synthetic accessibility (SA) score, molecular weight (MW), and novelty (fraction not present in the M2OR training set). For cross-profile analysis, we compute pairwise Tanimoto similarity using Morgan fingerprints (radius 2, 2048 bits) and extract Murcko scaffolds.


\section{Results}
\label{sec:results}

\subsection{Panoramic Binding Landscape}
\label{sec:results_panoramic}

The panoramic binding matrix reveals a highly structured landscape (\cref{fig:heatmap} in Appendix). Clustering identifies 8 distinct activation patterns, ranging from Cluster~1 (3{,}548 inert molecules binding no receptors) to Cluster~7 (105 broadly active molecules binding nearly all receptors).

Notably, Clusters~4 and~5 exhibit complementary selectivity with non-overlapping receptor sets, suggesting natural orthogonality in the olfactory code. The binding rates vary substantially across receptors: some receptors (e.g., s\_449) are activated by over 30\% of sampled molecules, while others (e.g., s\_1107) respond to fewer than 1\%, reflecting the heterogeneous promiscuity of the OR repertoire.

\subsection{Profile-Targeted Generation}
\label{sec:results_profiles}

\begin{figure}[htbp]
  \centering
  \includegraphics[width=\columnwidth]{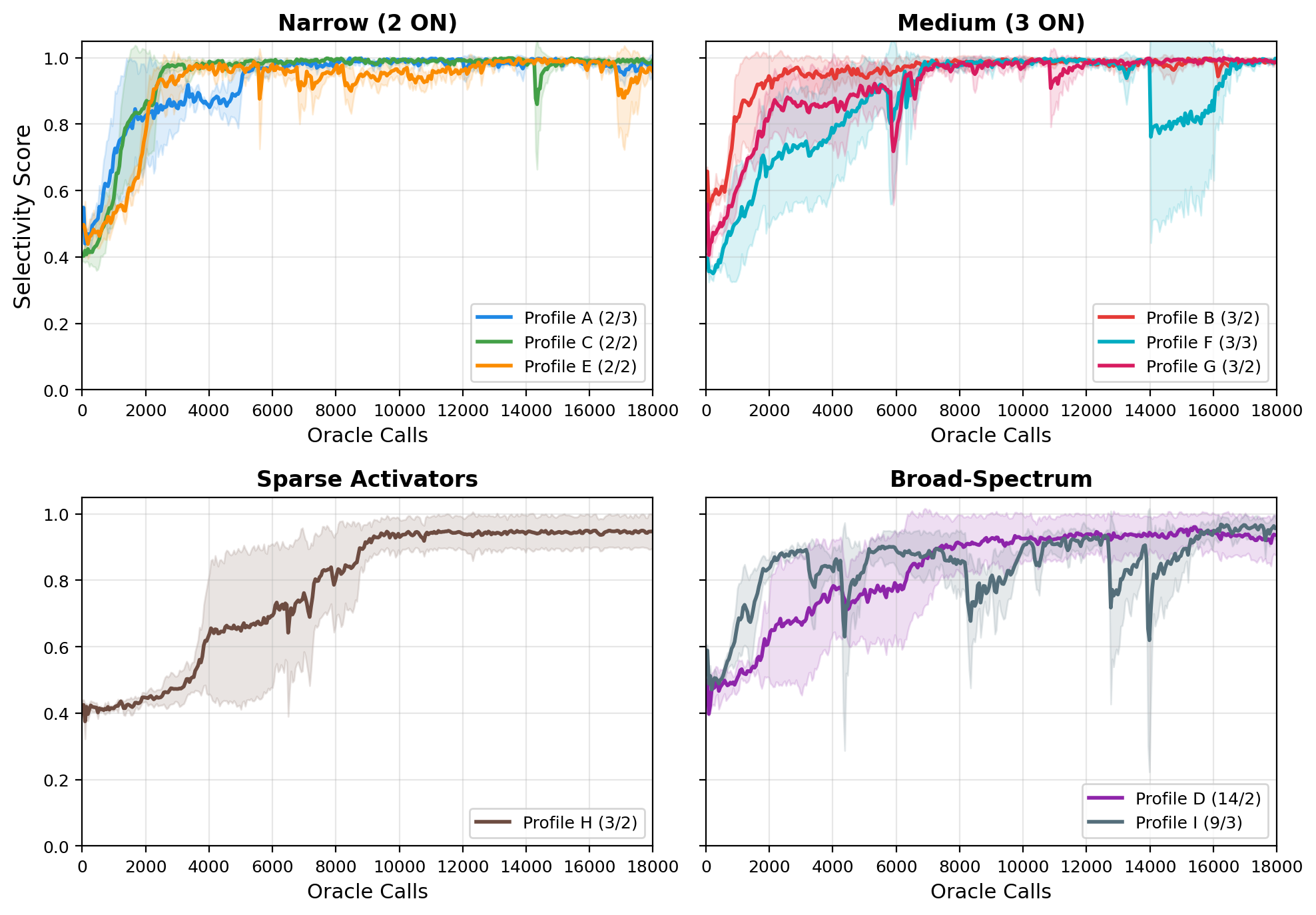}
  \caption{Training curves (structured loss, mean $\pm$ std, 3 seeds) grouped by profile type. \textbf{(a)} Narrow profiles (2 ON) converge almost instantly. \textbf{(b)} Medium profiles (3 ON) converge within $\sim$100 calls. \textbf{(c)} Profile H (sparse activators) requires $\sim$5{,}000 calls due to low-promiscuity ON receptors. \textbf{(d)} Broad-spectrum profiles D (14 ON) and I (9 ON) show the slowest convergence.}
  \label{fig:training_curves}
\end{figure}

\cref{fig:training_curves} shows the training dynamics for all nine profiles across 3 seeds. The selectivity score (mean over each batch) increases overall for all profiles, with convergence rate inversely related to profile complexity.

\begin{table*}[t]
  \caption{(\textbf{Top}) Target profile definitions: {\color{green!60!black}\checkmark} = ON receptor (target $=1$), {\color{red}\ding{55}} = OFF receptor (target $=0$), -- = unconstrained. (\textbf{Bottom}) Method comparison: each cell reports the peak batch-mean selectivity (mean $\pm$ std over 3 seeds, 18{,}000 oracle calls). Best result per profile in \textbf{bold}. \colorbox{green!8}{Profile D} and \colorbox{blue!6}{Profile H} are the most challenging targets.}
  \label{tab:results}
  \begin{center}
      \footnotesize
      \begin{tabular*}{\textwidth}{@{\extracolsep{\fill}}l*{20}{c}}
        \toprule
        & \rotatebox{90}{\tiny 29} & \rotatebox{90}{\tiny 51} & \rotatebox{90}{\tiny 111} & \rotatebox{90}{\tiny 125} & \rotatebox{90}{\tiny 177} & \rotatebox{90}{\tiny 272} & \rotatebox{90}{\tiny 353} & \rotatebox{90}{\tiny 405} & \rotatebox{90}{\tiny 412} & \rotatebox{90}{\tiny 418} & \rotatebox{90}{\tiny 443} & \rotatebox{90}{\tiny 449} & \rotatebox{90}{\tiny 464} & \rotatebox{90}{\tiny 482} & \rotatebox{90}{\tiny 539} & \rotatebox{90}{\tiny 949} & \rotatebox{90}{\tiny 950} & \rotatebox{90}{\tiny 1107} & \rotatebox{90}{\tiny 1116} & \rotatebox{90}{\tiny 1146} \\
        \midrule
        A (2/3) & {\color{green!60!black}\checkmark} & -- & -- & -- & -- & -- & -- & -- & -- & {\color{red}\ding{55}} & {\color{green!60!black}\checkmark} & -- & -- & -- & -- & {\color{red}\ding{55}} & {\color{red}\ding{55}} & -- & -- & -- \\
        B (3/2) & -- & {\color{green!60!black}\checkmark} & -- & -- & -- & -- & -- & -- & -- & -- & -- & {\color{green!60!black}\checkmark} & -- & -- & {\color{green!60!black}\checkmark} & {\color{red}\ding{55}} & {\color{red}\ding{55}} & -- & -- & -- \\
        C (2/2) & -- & {\color{red}\ding{55}} & -- & -- & -- & -- & -- & -- & -- & -- & -- & {\color{red}\ding{55}} & {\color{green!60!black}\checkmark} & {\color{green!60!black}\checkmark} & -- & -- & -- & -- & -- & -- \\
        \rowcolor{green!8} D (14/2) & {\color{green!60!black}\checkmark} & {\color{green!60!black}\checkmark} & -- & -- & {\color{green!60!black}\checkmark} & {\color{green!60!black}\checkmark} & {\color{green!60!black}\checkmark} & -- & {\color{green!60!black}\checkmark} & {\color{green!60!black}\checkmark} & {\color{green!60!black}\checkmark} & {\color{green!60!black}\checkmark} & {\color{red}\ding{55}} & {\color{green!60!black}\checkmark} & {\color{green!60!black}\checkmark} & {\color{green!60!black}\checkmark} & {\color{green!60!black}\checkmark} & {\color{red}\ding{55}} & {\color{green!60!black}\checkmark} & -- \\
        E (2/2) & {\color{red}\ding{55}} & {\color{green!60!black}\checkmark} & -- & -- & -- & -- & -- & -- & -- & -- & {\color{red}\ding{55}} & {\color{green!60!black}\checkmark} & -- & -- & -- & -- & -- & -- & -- & -- \\
        F (3/3) & {\color{green!60!black}\checkmark} & {\color{red}\ding{55}} & {\color{green!60!black}\checkmark} & -- & -- & -- & -- & -- & -- & -- & {\color{green!60!black}\checkmark} & {\color{red}\ding{55}} & -- & -- & {\color{red}\ding{55}} & -- & -- & -- & -- & -- \\
        G (3/2) & -- & -- & -- & {\color{green!60!black}\checkmark} & -- & -- & -- & -- & -- & -- & -- & -- & -- & {\color{green!60!black}\checkmark} & -- & {\color{red}\ding{55}} & {\color{red}\ding{55}} & -- & -- & {\color{green!60!black}\checkmark} \\
        \rowcolor{blue!6} H (3/2) & -- & -- & {\color{red}\ding{55}} & -- & -- & {\color{green!60!black}\checkmark} & -- & -- & -- & -- & {\color{red}\ding{55}} & -- & {\color{green!60!black}\checkmark} & -- & -- & -- & -- & -- & {\color{green!60!black}\checkmark} & -- \\
        I (9/3) & {\color{green!60!black}\checkmark} & {\color{green!60!black}\checkmark} & -- & {\color{green!60!black}\checkmark} & -- & -- & -- & {\color{green!60!black}\checkmark} & -- & -- & {\color{green!60!black}\checkmark} & {\color{green!60!black}\checkmark} & -- & {\color{green!60!black}\checkmark} & {\color{green!60!black}\checkmark} & {\color{red}\ding{55}} & {\color{red}\ding{55}} & {\color{red}\ding{55}} & -- & {\color{green!60!black}\checkmark} \\
        \bottomrule
      \end{tabular*}

      \vskip 0.1in

      \resizebox{\textwidth}{!}{%
      \begin{tabular}{l*{9}{c}}
        \toprule
        \textbf{\normalsize Method} & \textbf{A} & \textbf{B} & \textbf{C} & \cellcolor{green!8}\textbf{D} & \textbf{E} & \textbf{F} & \textbf{G} & \cellcolor{blue!6}\textbf{H} & \textbf{I} \\
        \midrule
        {\normalsize Random prior} & $\mathbf{1.00} \pm 0.00$ & $\mathbf{1.00} \pm 0.00$ & $0.99 \pm 0.00$ & \cellcolor{green!8} $0.92 \pm 0.04$ & $\mathbf{1.00} \pm 0.00$ & $0.99 \pm 0.01$ & $\mathbf{1.00} \pm 0.00$ & \cellcolor{blue!6} $0.83 \pm 0.00$ & $\mathbf{1.00} \pm 0.00$ \\
        {\normalsize Pan.\ screen.} & $\mathbf{1.00} \pm 0.00$ & $\mathbf{1.00} \pm 0.00$ & $0.93 \pm 0.01$ & \cellcolor{green!8} $0.88 \pm 0.03$ & $\mathbf{1.00} \pm 0.00$ & $0.97 \pm 0.00$ & $\mathbf{1.00} \pm 0.00$ & \cellcolor{blue!6} $0.80 \pm 0.01$ & $0.95 \pm 0.00$ \\
        \midrule
        {\normalsize W-MSE} & $\mathbf{1.00} \pm 0.00$ & $\mathbf{1.00} \pm 0.00$ & $\mathbf{1.00} \pm 0.00$ & \cellcolor{green!8} $0.87 \pm 0.16$ & $\mathbf{1.00} \pm 0.00$ & $\mathbf{1.00} \pm 0.00$ & $\mathbf{1.00} \pm 0.00$ & \cellcolor{blue!6} $0.92 \pm 0.07$ & $0.97 \pm 0.03$ \\
        {\normalsize Lagrangian} & $\mathbf{1.00} \pm 0.00$ & $\mathbf{1.00} \pm 0.00$ & $0.82 \pm 0.25$ & \cellcolor{green!8} $\mathbf{0.97} \pm 0.05$ & $\mathbf{1.00} \pm 0.00$ & $0.89 \pm 0.14$ & $0.99 \pm 0.00$ & \cellcolor{blue!6} $0.80 \pm 0.13$ & $0.92 \pm 0.06$ \\
        \textbf{\normalsize Structured} & $\mathbf{1.00} \pm 0.00$ & $\mathbf{1.00} \pm 0.00$ & $\mathbf{1.00} \pm 0.00$ & \cellcolor{green!8} $0.96 \pm 0.04$ & $\mathbf{1.00} \pm 0.00$ & $\mathbf{1.00} \pm 0.00$ & $\mathbf{1.00} \pm 0.00$ & \cellcolor{blue!6} $\mathbf{0.95} \pm 0.05$ & $0.99 \pm 0.01$ \\
        \bottomrule
      \end{tabular}}
  \end{center}
  \vskip -0.1in
\end{table*}

\cref{tab:results} presents the main results. Several patterns emerge:

\paragraph{Non-learning baselines are strong on easy profiles but weaker on rare targets.} For narrow profiles (A, B, E, G), both random prior sampling and panoramic screening achieve near-perfect selectivity, as suitable molecules are already well represented in the prior distribution. Random sampling also performs strongly on Profile I, likely because its ON set contains several promiscuous receptors. The clearest gaps appear for Profile D and especially Profile H: panoramic screening reaches $0.88$ and $0.80$, respectively, while the structured loss reaches $0.96$ and $0.95$. These results suggest that directed generation is most valuable when the target activation pattern is rare under the prior, particularly for sparse-activator profiles involving low-promiscuity receptors.

\paragraph{Structured loss improves broad-spectrum optimization.} The structured combinatorial code loss achieves the best or tied-best result on eight of nine profiles. On Profile D, it reaches $0.96 \pm 0.04$, comparable to the Lagrangian variant ($0.97 \pm 0.05$) but with lower variance, and substantially outperforming W-MSE ($0.87 \pm 0.16$). On Profile H, the hardest profile, it reaches $0.95 \pm 0.05$, outperforming both W-MSE ($0.92 \pm 0.07$) and the Lagrangian variant ($0.80 \pm 0.13$). On narrow profiles (A, B, E, G), all RL methods achieve $1.00$.

\begin{figure}[htbp]
  \centering
  \includegraphics[width=\columnwidth]{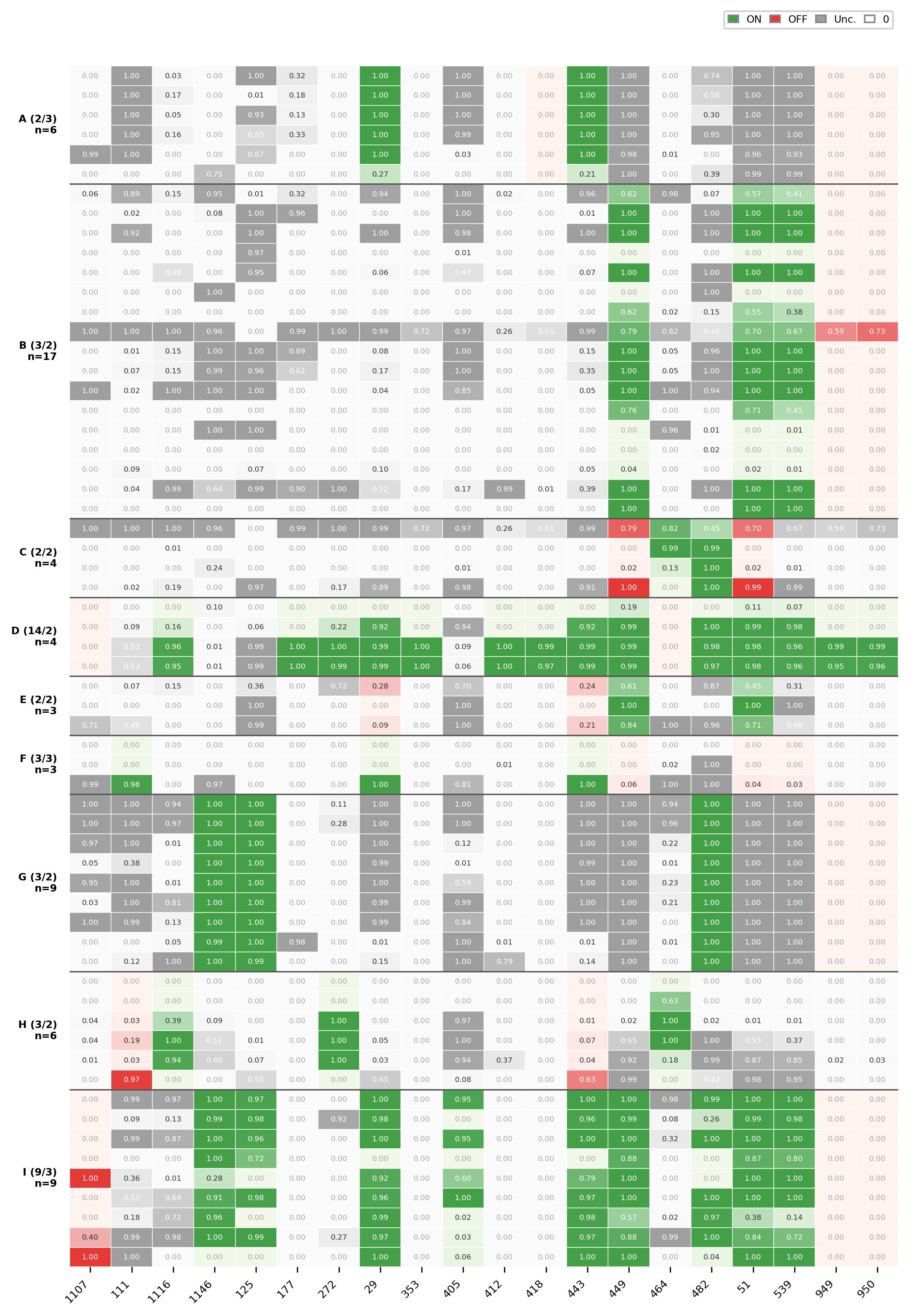}
  \caption{Per-receptor binding of generated molecules (structured loss, sel $> 0.8$, 3 seeds). Each row is one molecule, grouped by profile. Green = ON, red = OFF, gray = unconstrained. Intensity = binding probability.}
  \label{fig:per_receptor}
\end{figure}

\paragraph{Lagrangian constraints help simple selective profiles.} The constrained formulation achieves competitive results on simple profiles (A, B, E) but shows high variance on Profiles C ($0.825 \pm 0.247$), F ($0.894 \pm 0.145$), and I ($0.918 \pm 0.063$), suggesting that adaptive multipliers become unstable when many constraints must be balanced.

\paragraph{Designability gradient.} Optimization difficulty depends jointly on target dimensionality and ON-receptor promiscuity. Most narrow or medium profiles converge within $\sim$100 oracle calls. Profile I (9 ON) also converges quickly because its ON set contains promiscuous receptors. In contrast, Profile D requires $\sim$1{,}231 calls and Profile H requires $\sim$5{,}367 calls due to low baseline binding rates of its ON receptors. Broad profiles with promiscuous receptors can thus be easier than sparse profiles targeting low-promiscuity receptors.

\subsection{Cross-Profile Chemical Analysis}
\label{sec:results_chemical}

A critical question is whether molecules optimized for different activation patterns are chemically distinct. We collect the top-100 molecules from each profile and compute Morgan fingerprints (radius 2, 2048 bits). The mean inter-profile Tanimoto similarity ($0.120 \pm 0.105$) is slightly lower than intra-profile similarity ($0.128 \pm 0.100$), and scaffold overlap between any two profiles is below 35\%, suggesting partial profile-specific chemical separation.

\begin{table}[htbp]
  \caption{Quality metrics of top generated molecules per profile (structured loss, aggregated across 3 seeds).}
  \label{tab:quality}
  \begin{center}
    \resizebox{\columnwidth}{!}{%
    \begin{footnotesize}
      \begin{tabular}{lccccc}
        \toprule
        \textbf{Prof.} & \textbf{Novel\,$\uparrow$} & \textbf{QED\,$\uparrow$} & \textbf{SA\,$\downarrow$} & \textbf{MW} & \textbf{cLogP} \\
        \midrule
        A & 100\% & $0.56 \pm 0.08$ & $1.87 \pm 0.80$ & $137 \pm 31$ & $3.18 \pm 0.56$ \\
        B & 65\% & $0.58 \pm 0.12$ & $2.13 \pm 0.69$ & $137 \pm 50$ & $1.57 \pm 1.08$ \\
        C & 50\% & $0.55 \pm 0.14$ & $1.77 \pm 0.59$ & $128 \pm 46$ & $1.70 \pm 0.56$ \\
        D & 100\% & $0.52 \pm 0.10$ & $1.95 \pm 0.73$ & $146 \pm 41$ & $2.48 \pm 0.37$ \\
        E & 67\% & $0.54 \pm 0.06$ & $1.84 \pm 0.60$ & $153 \pm 21$ & $1.73 \pm 1.34$ \\
        F & 100\% & $0.50 \pm 0.04$ & $1.84 \pm 0.41$ & $110 \pm 24$ & $2.04 \pm 1.00$ \\
        G & 100\% & $0.57 \pm 0.05$ & $1.99 \pm 0.83$ & $141 \pm 12$ & $1.60 \pm 0.44$ \\
        H & 67\% & $0.42 \pm 0.12$ & $2.31 \pm 0.09$ & $115 \pm 43$ & $1.16 \pm 1.47$ \\
        I & 89\% & $0.48 \pm 0.07$ & $2.04 \pm 0.41$ & $156 \pm 24$ & $2.95 \pm 0.99$ \\
        \bottomrule
      \end{tabular}
    \end{footnotesize}}
  \end{center}
  \vskip -0.1in
\end{table}

QED scores remain in the chemically plausible range (0.4--0.6), SA scores ($<$2.6) indicate synthetic accessibility, and molecular weights (110--164 Da) are typical of volatile odorants. cLogP values (1.5--3.4) are consistent with membrane-permeable compounds. Novelty ranges from 50\% (Profile C) to 100\% (Profiles A, D, F, G). Some profiles show repeated high-scoring molecules across seeds, suggesting possible mode collapse, a limitation we discuss further below.

\subsection{Ablation: Reward Formulation}
\label{sec:results_ablation}

To understand how reward formulation interacts with profile complexity, we group the nine profiles into three difficulty tiers (Easy: A, B, E, G with 2--3 ON receptors and high baseline binding; Medium: C, F, I with cross-cluster or high-dimensional targets; Hard: D, H with broad-spectrum or sparse-activator targets) and compare the three RL reward formulations (\cref{tab:ablation}).

\begin{table}[htbp]
  \caption{Reward formulation comparison across profile difficulty groups (3 seeds per profile). \emph{Sel.}: mean best-of-budget selectivity (peak single-molecule score per seed). \emph{\%$\geq$0.99}: fraction of seeds reaching selectivity $\geq 0.99$. $\bar{\sigma}$: mean per-profile std computed from unrounded per-seed values (lower = more stable). \emph{\#Solved}: profiles with mean selectivity $\geq 0.99$.}
  \label{tab:ablation}
  \begin{center}
    \resizebox{\columnwidth}{!}{%
    \begin{tabular}{lcccccc}
      \toprule
      & \multicolumn{2}{c}{\textbf{Easy (4)}} & \multicolumn{2}{c}{\textbf{Hard (2)}} & & \\
      \cmidrule(lr){2-3} \cmidrule(lr){4-5}
      \textbf{Reward} & Sel.\,$\uparrow$ & \%$\geq$0.99\,$\uparrow$ & Sel.\,$\uparrow$ & \%$\geq$0.99\,$\uparrow$ & $\bar{\sigma}$\,$\downarrow$ & \#Solved\,$\uparrow$ \\
      \midrule
      W-MSE & $1.00$ & 100\% & $0.95$ & 50\% & $0.013$ & 7/9 \\
      Lagrangian & $1.00$ & 100\% & $0.91$ & 33\% & $0.021$ & 6/9 \\
      \textbf{Structured} & $\mathbf{1.00}$ & \textbf{100\%} & $\mathbf{0.99}$ & \textbf{67\%} & $\mathbf{0.002}$ & \textbf{8/9} \\
      \bottomrule
    \end{tabular}}
  \end{center}
  \vskip -0.1in
\end{table}

On easy profiles, all three formulations achieve perfect selectivity with 100\% of seeds reaching $\geq 0.99$, since the optimization landscape is simple enough that any reasonable reward signal suffices. The differences emerge on hard profiles, where the structured loss maintains $0.99$ selectivity with 67\% of seeds reaching $\geq 0.99$, compared to 50\% for W-MSE ($0.95$ mean) and only 33\% for the Lagrangian variant ($0.91$ mean). Overall, the structured loss solves 8 of 9 profiles (mean selectivity $\geq 0.99$), versus 7 for W-MSE and 6 for Lagrangian. It is also the most stable across seeds ($\bar{\sigma} = 0.002$), 6--10$\times$ lower than alternatives.

We attribute this robustness to two properties of the structured formulation. First, the pairwise ranking component (\cref{eq:ranking_loss}) provides gradient signal based on \emph{relative} ON--OFF ordering rather than absolute binding values, making it robust to calibration error in the binding predictor. Second, the distribution matching component (\cref{eq:js}) captures the full response geometry as a distribution, rather than treating each receptor as an independent regression target as W-MSE does. Together, these provide a smoother optimization landscape that scales to high-dimensional profiles.

\begin{figure*}[htbp]
  \centering
  \includegraphics[width=\textwidth]{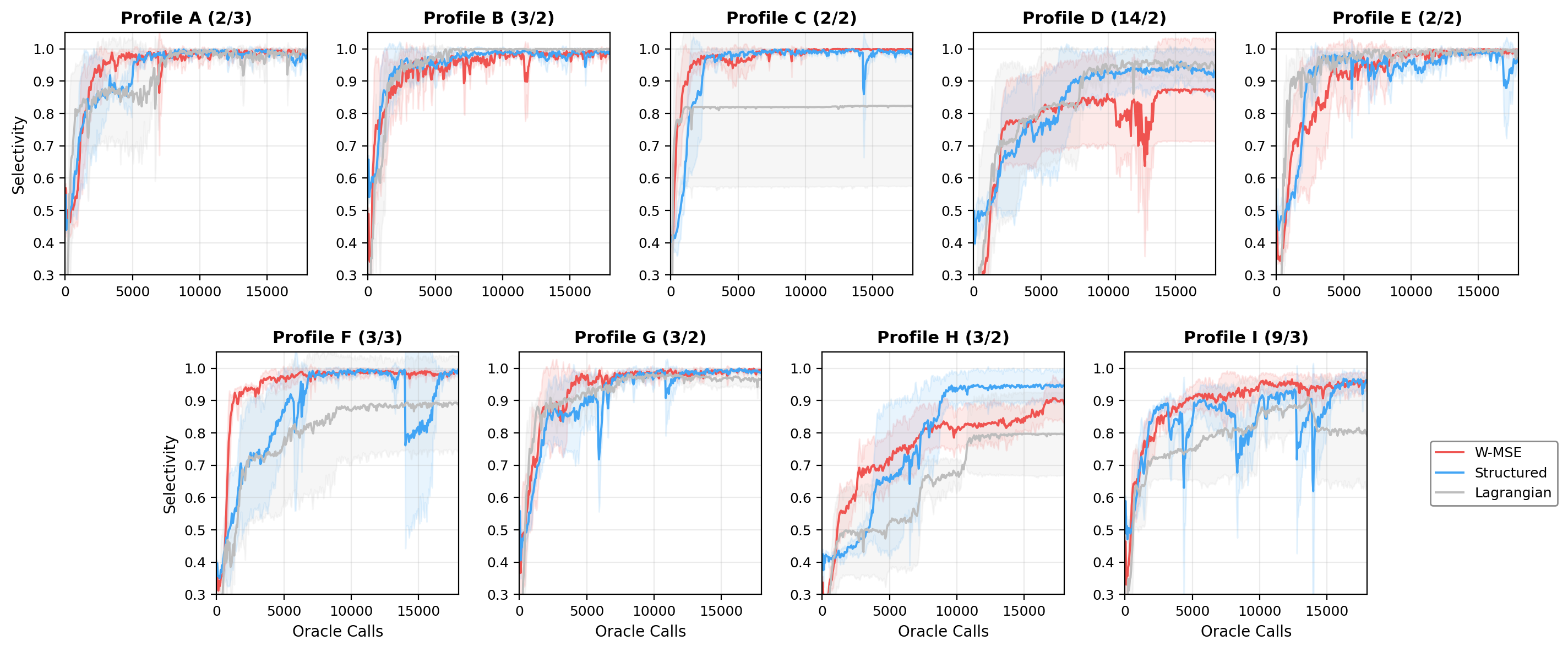}
  \caption{Training curves comparing three reward formulations across all nine profiles (mean $\pm$ std shaded, 3 seeds each). The structured loss matches or outperforms W-MSE on all profiles, with the largest improvement on Profiles D and H. The Lagrangian variant shows high variance on Profiles C, F, and I.}
  \label{fig:method_comparison}
\end{figure*}

\cref{fig:method_comparison} compares the training dynamics across all nine profiles. The structured loss consistently reaches higher selectivity than W-MSE, particularly on Profile D ($0.96$ vs.\ $0.87$). The Lagrangian constrained variant shows instability on several profiles (C, F, I), where balancing multiple ON/OFF constraints with adaptive multipliers proves challenging.

\section{Discussion}
\label{sec:discussion}

\paragraph{Structured loss vs.\ heuristic reward.}
A key design insight is that the structured combinatorial code loss (\cref{eq:combined}) provides a richer and more principled optimization signal than simple weighted MSE. The ranking component (\cref{eq:ranking_loss}) directly encodes the selectivity requirement, that ON receptors should outrank OFF receptors, without requiring accurate absolute calibration of the binding predictor. The distribution matching component (\cref{eq:js}) captures the full response geometry, treating the receptor profile as a distribution rather than independent binary targets. The complexity term (\cref{eq:complexity}) ensures that the generator matches not just which receptors are active, but the shape of the activation pattern. Together, these components provide stable gradient signal from the start of training: even a molecule that partially satisfies the ranking constraints receives meaningful reward.

\paragraph{When is RL-based generation necessary?}
Our baseline comparison (\cref{tab:results}) reveals that for narrow profiles (A, B, E, G), suitable molecules already exist in the prior distribution, and panoramic screening achieves near-perfect selectivity. However, for Profile D and especially Profile H, the gap between screening and RL-based generation is substantial. On Profile D, the structured loss achieves $0.96$ vs.\ $0.88$ for panoramic screening. On Profile H, the gap is $0.95$ vs.\ $0.80$. This suggests that RL-based generation is most valuable when the target activation pattern is rare under the prior.

\paragraph{Profile complexity and designability.}
Our results across nine profiles reveal that optimization difficulty depends jointly on target dimensionality and receptor promiscuity. Narrow profiles (2--3 ON receptors) converge to high selectivity within $\sim$100 calls, while Profile D (14 ON) requires $\sim$1{,}231 and Profile H (sparse activators) requires $\sim$5{,}367. Notably, Profile H is harder than Profile D despite having fewer receptors, because its ON receptors (s\_1116, s\_272, s\_464) have low baseline binding rates ($<$5\%). This ``designability gradient'' has practical implications: designability is not determined solely by the number of ON receptors, and broad profiles containing promiscuous receptors can be relatively easy, whereas sparse profiles targeting low-promiscuity receptors can be substantially harder.

\paragraph{Chemical space structure.}
The cross-profile chemical analysis (\cref{sec:results_chemical}) suggests partial, but not complete, profile-specific structure in chemical space. The mean inter-profile Tanimoto similarity is only slightly lower than the mean intra-profile similarity, indicating that fingerprint-level separation is modest. However, scaffold overlap between profile pairs remains below 35\%, suggesting that different target profiles can enrich for partially distinct scaffold families. We therefore interpret these results as evidence of profile-dependent chemical enrichment rather than strong separation into disjoint chemical regions.

\paragraph{Limitations.}
Our framework relies entirely on computational binding predictions, which may have systematic biases for receptors with limited training data or sparse negative labels (mean AUROC varies from $0.58$ to $0.94$). Generated molecules achieving high predicted selectivity require experimental validation, as some may exploit predictor artifacts rather than genuine binding properties. Our best-of-budget metric evaluates discovery of at least one high-scoring molecule rather than average sample quality or diversity. The RL setup can collapse toward a small number of high-scoring scaffolds for some profiles, and our current approach trains a separate generator per profile rather than a single conditional model. Finally, the panoramic profiling stage is limited to 20 receptors, and scaling to the full human OR repertoire ($\sim$400 receptors) would require more efficient prediction methods.

\paragraph{Future work.}
Natural extensions include a profile-conditioned generator that takes the target activation pattern as input, enabling single-model generation across all profiles without retraining. Uncertainty-aware optimization using ensemble predictors could address surrogate bias by down-weighting molecules outside the predictor's reliable domain. Experimental validation of top-scoring molecules via calcium imaging assays would confirm whether predicted selectivity translates to genuine receptor activation. Finally, incorporating perceptual models that map receptor activation patterns to predicted odor qualities would close the loop from molecular design to sensory experience.

\section*{Acknowledgments}
This work was supported by the National Natural Science Foundation of China (Grant No. 52505043), the National Key Research and Development Program of China (Grant No. 2024YFB3212300), and the Fundamental Research Funds for the Central Universities (Grant No. 226-2025-00037). The authors thank Liujing Zhuang from the College of Biomedical Engineering and Instrument Science, Zhejiang University, and Hao Jia from the Shanghai Institute of Microsystem and Information Technology, Chinese Academy of Sciences, for their valuable discussions in the early stages of this work.

\section*{Impact Statement}

This paper presents a computational framework for designing molecules with targeted olfactory receptor activation patterns. Potential applications include fragrance design and flavor chemistry. As with all molecular generation methods, generated candidates require safety and toxicity screening before any real-world use. We do not foresee specific negative societal consequences beyond those common to the field of de novo molecular design.

\bibliography{references}
\bibliographystyle{icml2026}

\newpage
\appendix
\onecolumn

\section{Target Profile Definitions}
\label{app:profiles}

The nine target profiles are derived from the cluster analysis of the panoramic binding matrix. Each profile specifies a set of ON receptors (target $t_i = 1$) and OFF receptors (target $t_i = 0$), with the remaining receptors assigned weight $w_i = 0$ (``don't care''). The profiles are designed to span a range of complexities.

\textbf{Profile A} targets the activation pattern of Cluster~5 (s\_29/s\_443 selective) while suppressing Cluster~4 receptors (s\_949/s\_950/s\_418). These two clusters have non-overlapping receptor sets, making this an ``orthogonal'' selectivity task. \textbf{Profile B} targets Cluster~6 (s\_449/s\_51/s\_539) against the same Cluster~4 OFF receptors. With 3 ON receptors, this tests whether the framework scales to slightly more complex patterns. \textbf{Profile C} targets Cluster~0 (s\_464/s\_482) while suppressing Cluster~6 receptors (s\_449/s\_51), a ``cross-cluster'' task where the ON and OFF receptor sets come from clusters with partial overlap in the panoramic matrix. \textbf{Profile D} targets the broad-spectrum pattern of Cluster~7, requiring activation of 14 receptors simultaneously while suppressing 2, testing the limits of the framework on high-dimensional profiles. \textbf{Profile E} targets an anti-correlated pair: ON receptors s\_449/s\_51 vs.\ OFF receptors s\_29/s\_443. Within each pair, receptors are highly correlated ($r > 0.99$), but the two pairs are uncorrelated, testing fine-grained selectivity. \textbf{Profile F} tests group-level selectivity: 3 ON (s\_111, s\_29, s\_443) vs.\ 3 OFF (s\_449, s\_51, s\_539), requiring simultaneous activation and suppression of equal-sized receptor groups. \textbf{Profile G} targets 3 ON (s\_1146, s\_125, s\_482) vs.\ 2 OFF (s\_949, s\_950) with moderate receptor promiscuity, serving as a control for medium-complexity profiles. \textbf{Profile H} targets sparse activators: 3 ON receptors (s\_1116, s\_272, s\_464) with low baseline binding rates ($<$5\%), making this the hardest optimization target despite having few receptors. \textbf{Profile I} targets a very broad pattern: 9 ON / 3 OFF receptors, second only to Profile~D in dimensionality.

ON receptors are weighted at $w_i = 2$ and OFF receptors at $w_i = 1$, reflecting the asymmetry that activating targets is typically harder than suppressing off-targets. Profile D uses uniform weights $w_i = 1$ for all active receptors to avoid over-penalizing any single receptor in the large ON set.

\begin{table}[h]
  \caption{Detailed target profile specifications.}
  \label{tab:profiles}
  \begin{center}
    \begin{small}
      \begin{tabular}{llll}
        \toprule
        \textbf{Profile} & \textbf{ON Receptors} & \textbf{OFF Receptors} & \textbf{Source} \\
        \midrule
        A & s\_29, s\_443 & s\_418, s\_949, s\_950 & C5 vs C4 \\
        B & s\_449, s\_51, s\_539 & s\_949, s\_950 & C6 vs C4 \\
        C & s\_464, s\_482 & s\_449, s\_51 & C0 selective \\
        D & s\_1116, s\_177, s\_272, s\_29, s\_353, s\_412, & s\_1107, s\_464 & C7 broad \\
          & s\_418, s\_443, s\_449, s\_482, s\_51, s\_539, s\_949, s\_950 & & \\
        E & s\_449, s\_51 & s\_29, s\_443 & Anti-correlated \\
        F & s\_111, s\_29, s\_443 & s\_449, s\_51, s\_539 & Group vs group \\
        G & s\_1146, s\_125, s\_482 & s\_949, s\_950 & Mixed medium \\
        H & s\_1116, s\_272, s\_464 & s\_111, s\_443 & Sparse activators \\
        I & s\_1146, s\_125, s\_29, s\_405, s\_443, & s\_1107, s\_949, s\_950 & Very broad \\
          & s\_449, s\_482, s\_51, s\_539 & & \\
        \bottomrule
      \end{tabular}
    \end{small}
  \end{center}
\end{table}

\section{Ablation Details}
\label{app:ablation}

The ablation in \cref{tab:ablation} groups nine profiles into three difficulty tiers based on the number of constrained receptors and their baseline binding rates. \textbf{Easy} profiles (A, B, E, G) involve 2--3 ON receptors with moderate-to-high baseline binding rates; suitable molecules already exist in the prior distribution and all RL methods converge within $\sim$100 oracle calls. \textbf{Medium} profiles (C, F, I) involve cross-cluster selectivity or high-dimensional targets (up to 9 ON receptors); these require the generator to explore beyond the prior but remain tractable for all methods. \textbf{Hard} profiles (D, H) represent the most challenging targets: Profile D requires simultaneous activation of 14 receptors, while Profile H targets 3 low-promiscuity receptors with baseline binding rates $<$5\%. These profiles expose the largest differences between reward formulations.

For each method--profile combination, we report the maximum selectivity score achieved across the 18{,}000-call budget, averaged over 3 seeds. The success rate (\%$\geq$0.99) measures the fraction of seeds that reach near-perfect selectivity. The stability metric $\bar{\sigma}$ is the mean per-profile standard deviation across seeds, measuring how consistently a method performs across random initializations.

\section{Panoramic Binding Matrix}
\label{app:heatmap}

\begin{figure}[h]
  \centering
  \includegraphics[width=\textwidth]{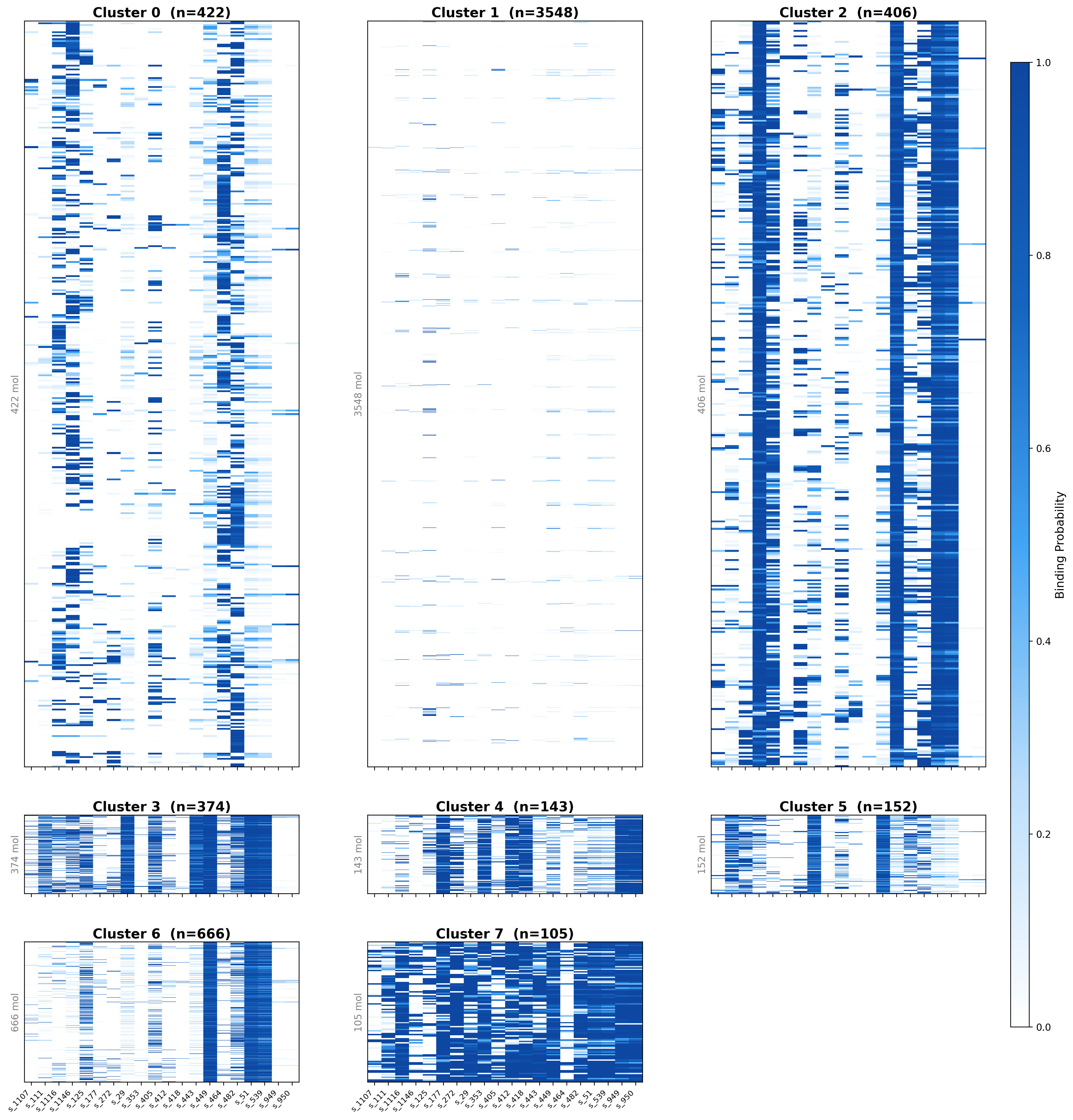}
  \caption{Panoramic binding matrix (5{,}816 molecules $\times$ 20 receptors), split by cluster. Color intensity indicates predicted binding probability. Eight clusters emerge with distinct activation patterns, from inert (C1) to broadly active (C7).}
  \label{fig:heatmap}
\end{figure}

\cref{fig:heatmap} shows the full panoramic binding matrix, with molecules sorted by cluster assignment. Each panel corresponds to one of the 8 discovered clusters. \textbf{Cluster 1} (n=3{,}548) is the dominant cluster, comprising 61\% of all sampled molecules with near-zero binding probability across all 20 receptors, representing the ``inert'' background of chemical space. \textbf{Clusters 2, 3, and 6} share a common set of active receptors (s\_449, s\_51, s\_539) but differ in the breadth and intensity of activation; Cluster 3 additionally activates s\_29 and s\_443, while Cluster 2 activates s\_1146. \textbf{Cluster 4} (n=143) has a distinctive pattern centered on s\_949, s\_950, s\_177, and s\_418, receptors that are largely inactive in other clusters, making Cluster 4 an ideal OFF-target set for Profiles A and B. \textbf{Cluster 5} (n=152) is selective for s\_29 and s\_443 with moderate activation of s\_111 and s\_1116, complementary to Cluster 4 and motivating Profile A. \textbf{Cluster 7} (n=105) is the broadly active cluster, with high binding probability across nearly all receptors, motivating Profile D. \textbf{Cluster 0} (n=422) shows moderate activation of s\_464 and s\_482, receptors that are relatively quiet in other clusters, motivating Profile C.

The per-receptor binding rates vary from 0.9\% (s\_1107) to 32\% (s\_449), reflecting the heterogeneous promiscuity of the OR panel. This variation is biologically meaningful: some receptors are broadly tuned ``generalists'' while others are narrowly tuned ``specialists.''

\section{Top Generated Molecules}
\label{app:top_molecules}

\begin{figure}[h]
  \centering
  \includegraphics[width=\textwidth]{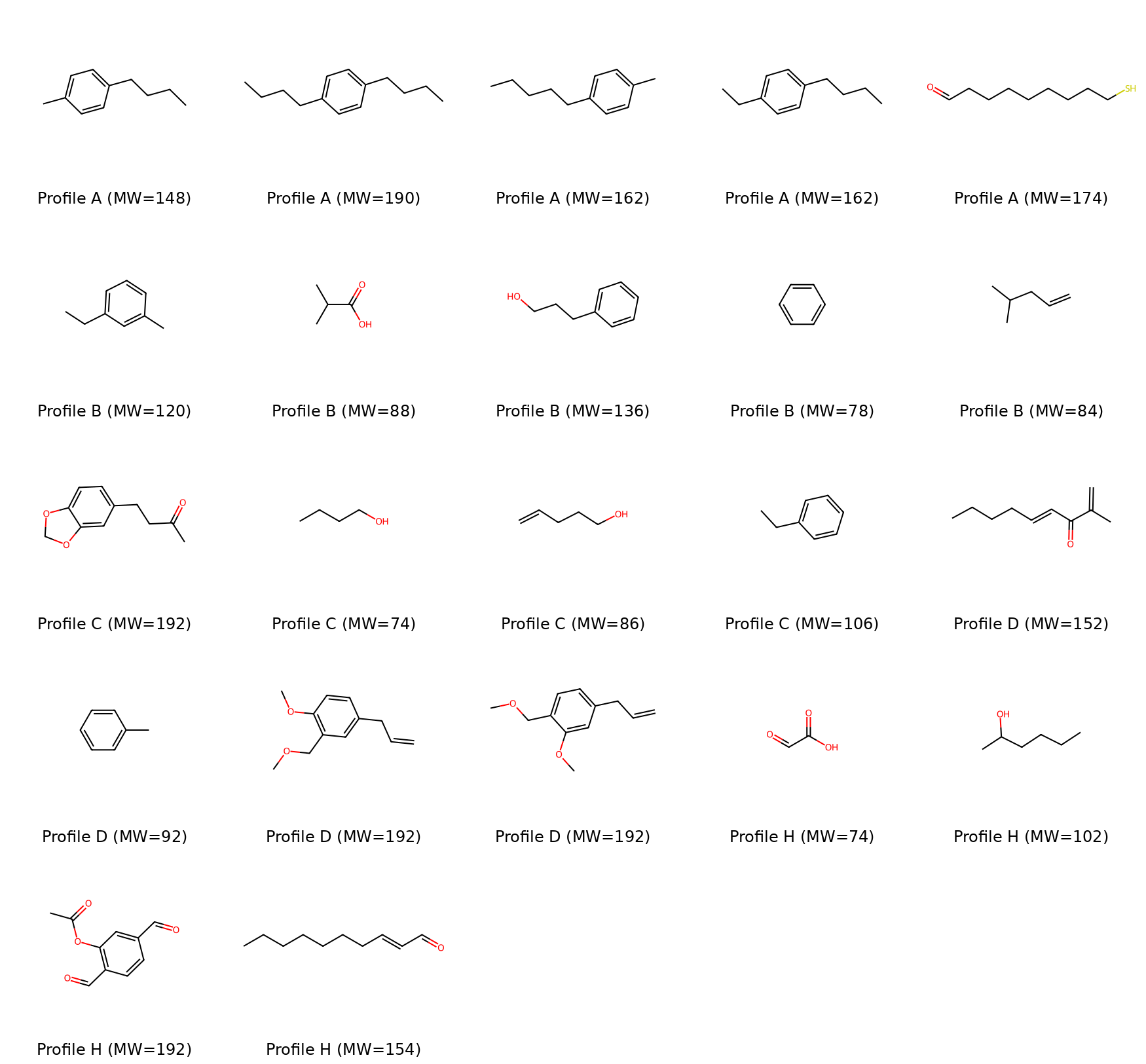}
  \caption{2D structures of top unique generated molecules for representative profiles (rows: A, B, C, D, H), filtered for odorant-likeness (MW $\leq$ 200). Molecules are selected by highest selectivity reward across 3 seeds.}
  \label{fig:top_molecules}
\end{figure}

\cref{fig:top_molecules} shows the 2D structures of the top generated molecules for representative profiles, filtered for odorant-likeness (MW $\leq$ 200, $\geq$ 5 heavy atoms, no charged species or nitro groups, at least 3 carbon atoms).

\paragraph{Profile A} molecules are predominantly linear or branched aliphatic chains with aldehyde, ketone, or ester functional groups (MW 114--158). This is consistent with known s\_29/s\_443 ligands, which tend to be medium-chain aldehydes and ketones.

\paragraph{Profile B} molecules show more structural diversity, including a thioaldehyde (O=CCCCS), isobutyraldehyde, menthol-like cyclohexanol, and hexanal. The presence of sulfur-containing compounds is notable and may reflect the binding preferences of s\_449/s\_51/s\_539.

\paragraph{Profile C} has fewer unique molecules (4 after filtering), reflecting the narrower chemical space that satisfies the s\_464/s\_482 selectivity constraint. The molecules include a methylenedioxyphenyl ketone, small ethers, and diols.

\paragraph{Profile D} molecules include aromatic aldehydes (naphthaldehyde), terpene esters (geranyl acetate, linalyl acetate), and a thiophenol derivative. The structural diversity is higher than for narrow profiles, consistent with the broad-spectrum activation requirement spanning many receptor types.

\paragraph{Profile H} molecules tend toward smaller, polar structures consistent with the sparse-activator receptor set. The low baseline binding rates of ON receptors (s\_272, s\_464, s\_1116) constrain the accessible chemical space.

All generated molecules have MW in the 72--196 Da range typical of volatile odorants, SA scores of 1.7--2.4 (synthetically accessible), and QED scores of 0.4--0.6, used here as a generic chemical plausibility filter rather than an odorant-specific objective. Novelty ranges from 50\% to 100\% across profiles.

\end{document}